\documentclass{bmvc2k}

\title{Motion and Context-Aware Audio-Visual Conditioned Video Prediction}

\addauthor{Yating Xu}{xu.yating@u.nus.edu}{1}
\addauthor{Conghui Hu}{conghui@nus.edu.sg}{1}
\addauthor{Gim Hee Lee}{gimhee.lee@nus.edu.sg}{1}

\addinstitution{
 Department of Computer Science \\
 National University of Singapore\\
 Singapore
}

\runninghead{Xu \etal}{Audio-Visual Conditioned Video Prediction}

\def\etal{\emph{et al}\bmvaOneDot}
\newcommand{\ie}{\textit{i.e.}~}

\usepackage{amsmath,amssymb} 
\usepackage{color}
\usepackage{booktabs}
\usepackage{multirow}
\usepackage{tabularx,verbatim}
\usepackage{xspace}
\usepackage{algorithm}
\usepackage{algorithmic}
\usepackage{setspace}
\usepackage{comment}
\usepackage{bbm}
\usepackage{enumitem}
\usepackage{colortbl}
\usepackage{color, xcolor} 
\usepackage[T1]{fontenc}
\usepackage{bbm}
\usepackage{wrapfig}
\newcommand{\ytComment}[1]{\textcolor{black}{#1}}
\begin{document}

\maketitle

\begin{abstract}
The existing state-of-the-art method for audio-visual conditioned video prediction 
uses the latent codes of the audio-visual frames from a multimodal stochastic network and a frame encoder to predict the next visual frame. However, a direct inference of per-pixel intensity for the next visual frame is extremely challenging because of the high-dimensional image space. 
To this end, we decouple the audio-visual conditioned video prediction into motion and appearance modeling. The multimodal motion estimation predicts future optical flow based on the audio-motion correlation. The visual branch recalls from the motion memory built from the audio features to enable better long-term prediction. 
We further propose context-aware refinement to address the diminishing of the global appearance context in the long-term continuous warping. The global appearance context is extracted by the context encoder and manipulated by motion-conditioned affine transformation before fusion with features of warped frames. 
Experimental results show that our method achieves competitive results on existing benchmarks.
\end{abstract}


\begin{wrapfigure}{R}{0.48\textwidth}
\vspace{-5mm}
\centering
\setlength{\abovecaptionskip}{0.1cm}
\includegraphics[scale=0.52]{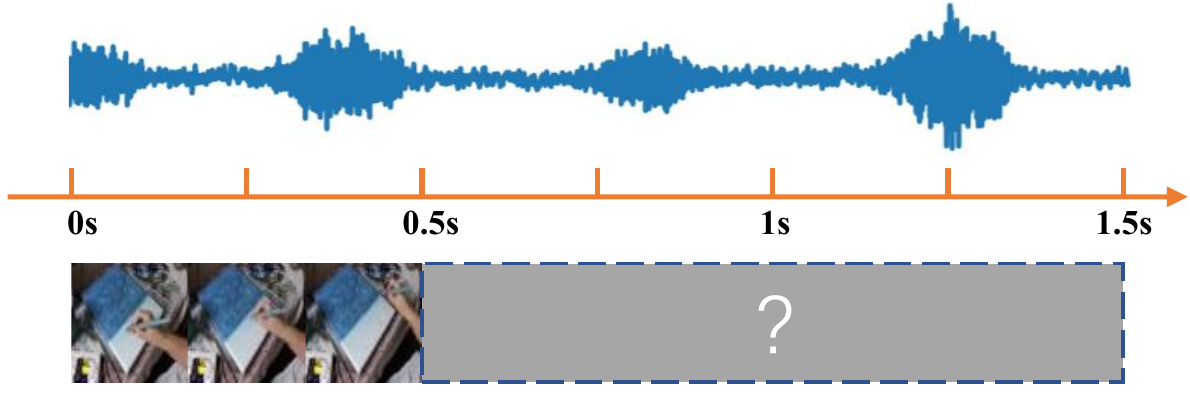}
\vspace{-3mm}
\caption{An illustration of audio-visual conditioned video prediction task. Given a full audio clip and a short sequence of past visual frames, it aims to predict corresponding future visual frames.}
\label{task intro}
\vspace{-3mm}
\end{wrapfigure}

\section{Introduction}
Humans perceive the world via multisensory processing. Particularly, audio-visual interactions are ubiquitous in our daily life. For instance, a person at a bus stop perceives an approaching bus through both audio and visual cues, yet his/her view is suddenly blocked by another pedestrian. Despite this visual barrier, the person's cognitive system can still fill in the missing visual content using previously seen frames and unbroken audio information. 
It has also been proven in neurobiology studies that such audio-visual pairing has the advantage of increasing the activation of neurons \cite{knopfel2019audio} to enhance perceptual ability. Inspired by these observations, our aim is to investigate the feasibility of equipping machines with the capacity to leverage audio-visual interaction, employing audio-visual conditioned video prediction as a testbed.

    

Fig.~\ref{task intro} illustrates the audio-visual conditioned video prediction task.
Given a full audio clip and a short sequence of the past visual frames, the objective is to predict the missing future visual frames that are as close to the ground truth frames as possible.
The given audio clips and visual frames act as guidance in predicting the motion and appearance of future visual frames. 

Existing state-of-the-art work Sound2Sight \cite{chatterjee2020sound2sight} proposes to directly infer per-pixel intensity, which contains both motion and appearance of future images, from the audio-visual inputs.
A multi-modal stochastic network is used to encode the audio-visual information into a latent space distribution. A random vector drawn from this latent space distribution is then concatenated with the embedding of the current visual frame to simultaneously predict the motion and appearance of objects for the next visual frame, which is extremely challenging due to the high-dimensional nature of image space. 
As a result, Sound2Sight struggles to preserve the visual content in the generated visual frames, \ie, the appearance of the digits in Fig.~\ref{mnist-img}. 
We postulate that this inability to retain the visual content
is caused by the tightly entangled motion and appearance modeling in Sound2Sight. 

To this end, we model the motion and appearance separately in the audio-visual conditioned video prediction. We propose multimodal motion estimation (MME) to predict future object motion in the form of optical flow by using the audio-motion correlation. The rhythm of the sound and the object motion are naturally aligned since the sound originates from object motion \cite{blokhintzev1946propagation, tijdeman1975propagation}. 
Inspired by this natural phenomenon, we store the audio features in the motion memory to guide the visual motion prediction. In contrast to the internal memory in the ConvLSTMs that selectively remembers the past memory, the motion memory keeps all the past memory, which provides a good supplement in long-term motion prediction.
To effectively bridge the audio-visual modality gap, we further design operator \textit{Condense} to condense the large-size motion memory into a compact representation and operator \textit{Recall} for the visual branch to use the audio motion context.

Although the next frame can now be generated by simply warping the current frame according to the predicted motion, the long-term continuous warping in the video prediction would result in the diminishing of the global appearance context. In order to effectively retain the appearance of objects, we creatively design a context-aware refinement (CAR) module. Other than just using a U-Net to refine the warped frames \cite{bei2021learning,pan2019video}, we design a context encoder to supplement the global appearance context by extracting the visual features from the given past frame. 
Since the appearance details of each future frame may vary due to object movement, we perform the motion-conditioned affine transformation on the global context feature before inserting it into the U-Net decoder for better adaptation of the motion variance at each time step.    


    
    
Our contributions in this paper are as follows:
\begin{itemize}
\setlength{\itemsep}{-1mm}
    \item We propose the multimodal motion estimation module to predict the optical flow of the next visual frame with the help of motion memory built from the audio features. 
    We design the $\textit{Condense}$ and $\textit{Recall}$ operators to help the visual branch effectively utilize the audio motion memory.
    \item We introduce the context-aware refinement module to mitigate the loss of global appearance context in the long-term warpings by using the context encoder and motion-conditioned affine transformation. 
    \item We conduct extensive experiments on both synthetic and real-world datasets to verify the effectiveness of our method.
\end{itemize}

\begin{figure*}[t]
\centering
\includegraphics[scale=0.4]{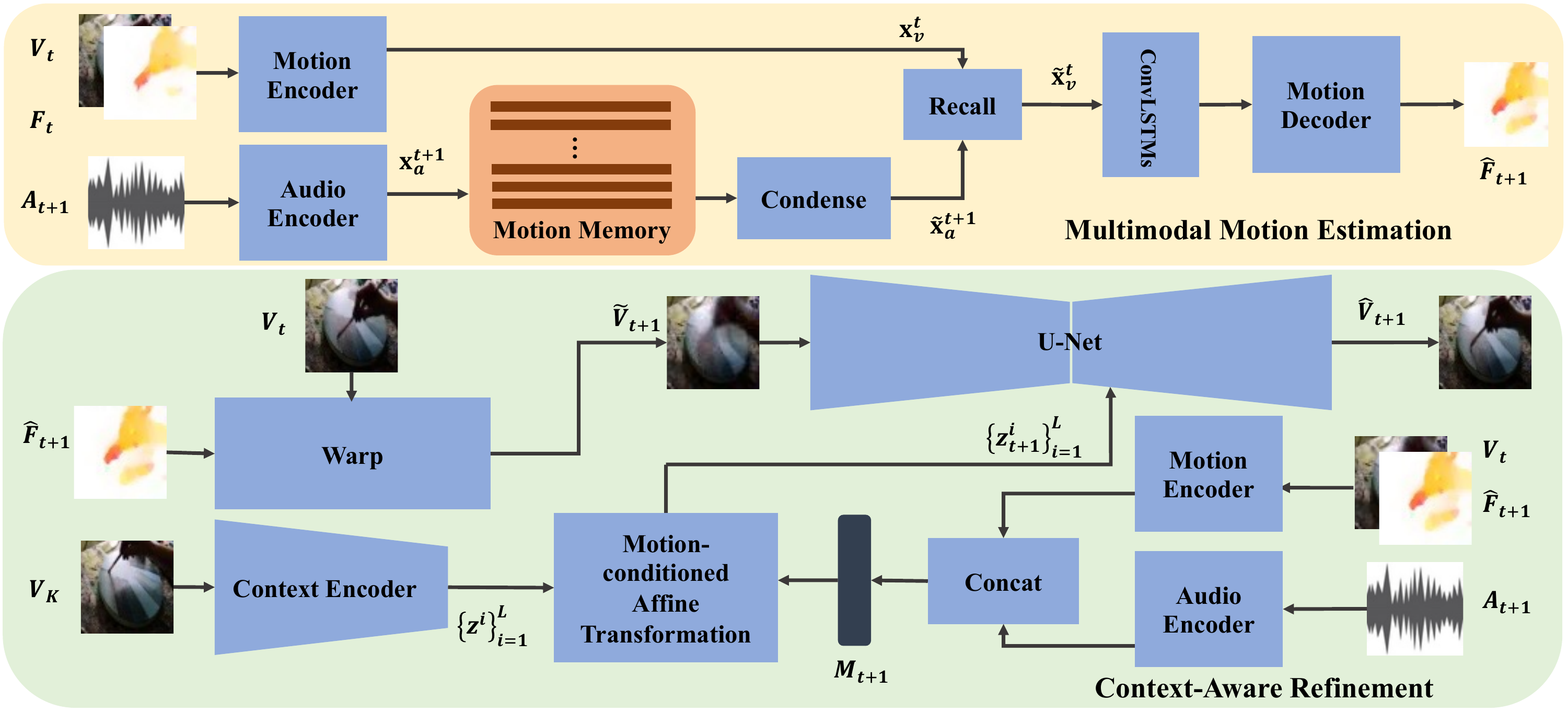}
\vspace{-3mm}
\caption{Our framework consists of the Multimodal Motion Estimation (MME) and Context-Aware Refinement (CAR) modules. MME predicts future optical flow by using audio information in the form of motion memory. CAR mitigates the loss of appearance context in the long-term recurrent warpings by using a context encoder and motion-conditioned affine transformation. 
}
\label{framework}
\vspace{-3mm}
\end{figure*}

\section{Related Work}
\noindent\textbf{Video prediction.}
Video prediction estimates future video frames close to ground truth future frames conditioned on a few past frames. It can be divided into two categories: prediction in pixel space \cite{wang2017predrnn, jin2020exploring, xingjian2015convolutional, lee2021video, wang2018eidetic}, and prediction in low-dimensional space \cite{lee2021revisiting, villegas2017learning, villegas2018hierarchical, wu2020future}. Prediction in pixel space directly estimates the per-pixel intensity. 
For example, PredRNN \cite{wang2017predrnn} enables memory states to be updated across vertically stacked RNN layers and horizontally through all RNN states. 
Jin \etal \cite{jin2020exploring} adopt multi-level wavelet transform to deal with motion blur and missing appearance details. 
Denton and Fergus \cite{denton2018stochastic} additionally consider stochasticity in the prediction by learning a prior distribution to capture the future dynamics.  

Direct estimation of the per-pixel intensity of future visual frames is a very challenging task because of the high-dimensional image space. As a result, there are many works that seek to predict a low-dimensional representation, such as semantic map \cite{lee2021revisiting, bei2021learning, villar2022MSPred}, human pose \cite{villegas2017learning, villar2022MSPred} and optical flow \cite{bei2021learning, li2018flow, liu2017video,liang2017dual,akan2021slamp}. Li \etal \cite{li2018flow} generate consecutive multiple future frames from only one image by multiple time step flow prediction and flow-to-frame synthesis. 
Bei \etal \cite{bei2021learning} first infer future optical flow and semantic maps independently for each object in the scene and then use U-Net to inpaint the future images. MSPred \cite{villar2022MSPred} simultaneously predicts future images and abstract representations via the hierarchical predictor module.



Different from the visual-only video prediction methods, we additionally consider the audio information. The pioneer work Sound2Sight \cite{chatterjee2020sound2sight}
learns a joint audio-visual latent embedding to capture future dynamics. However, the 
latent codes cause the tight entanglement of motion and appearance, thus making it difficult for the model to balance the capture of correct motion and maintaining the shape of objects. Our solution is to model the motion and appearance separately through multimodal motion estimation and context-aware refinement, respectively.


\noindent \textbf{Audio-visual representation learning.} Audio-Visual representation learning learns a general representation in a self-supervised way by utilizing audio-visual correlation. There are mainly three types of correspondences: semantic similarity \cite{arandjelovic2017look, aytar2016soundnet, arandjelovic2018objects,hu2019deep,morgado2021audio}, temporal consistency \cite{owens2018audio, korbar2018cooperative, afouras2020self} and spatial correspondence \cite{gao2020visualechoes, yang2020telling}. Arandjelovic 
 \etal \cite{arandjelovic2017look} propose audio-visual correspondence learning by classifying whether the input audio and visual frame are from the same video or not. \cite{owens2018audio} design a pretext task to predict whether the audio stream and visual sequence are temporally aligned. \cite{yang2020telling} leverage audio-visual spatial correspondence by predicting whether the two channels of audio are flipped. In addition to learning a general representation, there are also many downstream audio-visual tasks, such as sound source localization \cite{chen2021localizing, senocak2018learning, hu2020discriminative}, audio-visual source separation \cite{gao2019co, zhao2018sound, zhao2019sound}, audio-visual video parsing \cite{tian2020unified, wu2021exploring} and talking face generation \cite{chung2017you, chen2019hierarchical, vougioukas2018end}. In this paper, we study the task of audio-visual conditioned video prediction and leverage audio-motion correlation. 

\section{Method}
\begin{sloppypar}

\noindent\textbf{Problem definition.} Let us denote a video with $T$ frames as $\mathcal{V} = \{\{V_1, A_1\}, \cdots, \{V_T, A_T\} \}$, where $\{V_t, A_t\}$ represents a visual frame $V_t$ and its corresponding audio clip $A_t$. A visual frame $V_t$ is an image of size $C \times W \times H$, where $C$ is the number of channels, $W$ and $H$ are the width and height of the image.
An audio clip $A_t$ is a $\mathcal{F} \times \mathcal{T}$ two-dimensional spectrogram, where $\mathcal{F}$ denotes the frequency range and $\mathcal{T}$ represents the time duration of the audio clip. Given the first $K$ visual frames $\{V_1, \cdots, V_K\}$ and the whole audio $\{A_1, \cdots, A_K, \cdots, A_T\}$, the goal of the audio-visual conditioned video prediction task is to predict the missing $T-K$ visual frames $\{V_{K+1}, \cdots, V_T\}$ 
to be as similar to the ground truth visual frames as possible. For brevity, we drop the frame index $K$ and denote the indices of the previous, current and next predicted frames as $t-1$, $t$ and $t+1$, respectively. 
\end{sloppypar}

\vspace{2mm}
\noindent\textbf{Method overview.} Fig.~\ref{framework} shows our proposed framework which consists of two modules: 1) multimodal motion estimation (MME) and 2) context-aware refinement (CAR). Our MME and CAR modules are designed to model motion and appearance, respectively. 
MME predicts future optical flow by taking visual and audio inputs. 
It stores the past and current audio information in the motion memory and the visual branch recalls from the condensed motion memory to better infer long-term future motions. 
CAR renders future frames by taking global appearance context into account. The warped images are supplemented with global appearance features extracted by the context encoder, and modulated by motion-conditioned affine transformation to get the final predicted future frames.

\subsection{Multimodal Motion Estimation}
The MME predicts optical flow in a recurrent way. For ease of warping, the MME operates on backward optical flow, \ie both input $F_{t}$ and output $\hat{F}_{t+1}$ are backward optical flow. We use both audio and visual modalities in this module because they both contain clear motion information. To predict the $t+1$ frame, the motion encoder $E_m$ takes the concatenation of the current optical flow $F_{t}$ and visual frame $V_{t}$ as inputs, and outputs the motion feature $\text{x}_v^{t} \in \mathbb{R}^{w \times h \times c}$. Note that $V_{t}$ is a given visual frame if $t \leq K$, otherwise it is generated by warping $V_{t-1}$ according to the predicted optical flow $F_{t}$. We do not refine the warped image in MME. 
Concurrently, the audio encoder $E_a$ extracts the audio feature $\text{x}_a^{t+1} \in \mathbb{R}^{1 \times c}$ from input audio signal $A_{t+1}$. 

For our recurrent MME to effectively facilitate motion prediction, it is crucial to guarantee the presence of long-term memory for the audio information. However, the straightforward choice of LSTM is notorious for only having a short memory, where the memory cell is quickly saturated to remember only the latest input. As a result, it is difficult to infer long-term future motions. 
We thus design an external motion memory $\operatorname{MM}$ to store the audio features. In contrast to the memory cell inside LSTM that selectively remembers past memory, our motion memory $\operatorname{MM}$ keeps all the past memory without abandon. Our advantage is that it does not lose any past memory, and thus efficiently keeps all long-term information. The $\operatorname{MM}$ stores audio features up to time $t+1$, $\operatorname{MM}=\left\{\text{x}_{a}^n\right\}_{n=1}^{t+1} \in \mathbb{R}^{(t+1) \times c}$. 
Then, visual features \textit{Recall} from $\operatorname{MM}$ via cross-attention as the query $(\mathbf{Q})$, and $\operatorname{MM}$ is used to generate the key $(\mathbf{K})$ and value $(\mathbf{V})$. 
During the long-term prediction, the size of $\operatorname{MM}$ increases linearly with the prediction length, making it difficult for the visual branch to catch useful information from the large motion memory.  
We address this motion memory complexity problem by \textit{Condensing} the full $\operatorname{MM}$ into a compact memory $\tilde{\text{x}}_a^{t+1} \in \mathbb{R}^{1 \times c}$ as follows:
\ytComment{
\begin{equation}
    \begin{aligned}
    &\operatorname{Atten}(\mathbf{Q},\mathbf{K},\mathbf{V}) = \operatorname{Softmax}\left(\mathbf{Q} \mathbf{K}^\top\right) \mathbf{V}, \\
    \operatorname{Condense}(\operatorname{MM}) = &\operatorname{Atten}\left(\operatorname{MM}, \operatorname{MM}, \operatorname{MM}\right) [-1], \quad
    \tilde{\text{x}}_{a}^{t+1} = \text{x}_a^{t+1} + \operatorname{Condense}\left(\operatorname{MM}\right),\\
    \end{aligned}
\end{equation} 
}
%
where 
$[-1]$ denotes the last slice of the memory.
\ytComment{ 
The $\operatorname{Atten}$ operation follows \cite{wang2018non}.
}
The visual branch \textit{Recall}s from the condensed audio memory $\tilde{\text{x}}_a^{t+1}$ and obtain the enhanced motion feature $\tilde{\text{x}}_v^{t} \in \mathbb{R}^{w\times h \times c}$ as follow: 
\begin{equation}
    \operatorname{Recall}(\text{x}_v^{t}, \tilde{\text{x}}_a^{t+1}) 
    = \operatorname{Atten}(\text{x}_v^{t}, \tilde{\text{x}}_a^{t+1}, \tilde{\text{x}}_a^{t+1}), \quad
    \tilde{\text{x}}_v^{t} = \text{x}_v^{t} + \operatorname{Recall}(\text{x}_v^{t}, \tilde{\text{x}}_a^{t+1}).
\label{recall}
\end{equation}
We tile $x_v^t$ into the feature vector of size $l \times c$, where $l = w \times h $ before computing Eqn.~\ref{recall}. 
$\tilde{\text{x}}_v^{t}$ is then fed into ConvLSTMs and motion decoder $D_m$ to predict optical flow $\hat{F}_{t+1}$ at next time step as follows: 
\begin{equation}
    \{h_{t+1}, o_{t+1}\} = \operatorname{ConvLSTM}\left(\tilde{\text{x}}_v^{t},  h_{t}\right), \quad
    \hat{F}_{t+1} = D_m(h_{t+1}),
\end{equation} 
where $h_{t+1}$ is the hidden state and $o_{t+1}$ is the cell state of the ConvLSTMs at time step $t+1$. 


\begin{wrapfigure}{R}{0.48\textwidth}
\centering
\includegraphics[scale=0.43]{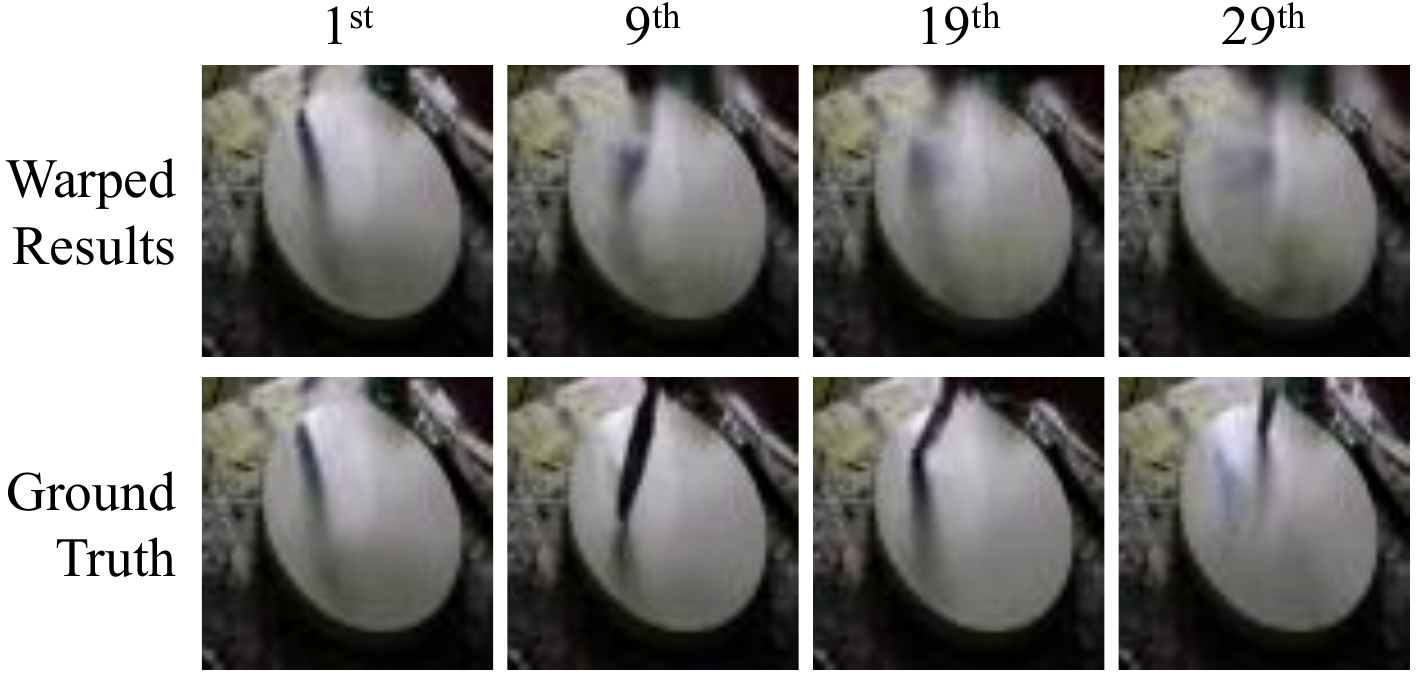}
\vspace{-8mm}
\caption{Example of the warped images. Resultant images gradually lose the shape of the brush.}
\label{warp}
\end{wrapfigure}

\subsection{Context-Aware Refinement}
The Context-Aware Refinement (CAR) module is then designed to generate the next visual frame according to the optical flow predicted by MME. 
Since audio cannot convey dense pixel information, we rely on the visual modality for appearance modeling. We start by warping the current visual frame $V_{t}$ using the predicted optical flow $\hat{F}_{t+1}$ as follow: 
\begin{equation}
\hat{V}_{t+1}\left(p\right) = V_{t}\left(p+\hat{F}_{t+1}(p)\right),
\end{equation}
where $p$ is the pixel coordinates. Similar to MME, $V_{t}$ is the given visual frame if $t \leq K$, otherwise it is the output of CAR.

The warping procedure is performed recurrently by taking the current prediction as input and warping it into the next frame. It causes the warped image to become increasingly blurry and consequently leads to the loss of appearance context as shown in Fig.~\ref{warp}. 
The warped images gradually lose the shape of the brush.

To alleviate the blurry visual frame problem, we first use a U-Net \cite{ronneberger2015u} to refine the local details. 
Furthermore, we address the loss of appearance context by providing external appearance information. The context encoder $E_c$ extracts global appearance context $Z=\left\{\text{z}^i\right\}_{i=1}^{L}$ from the last given visual frame $V_K$. $\text{z}^i \in \mathbb{R}^{c^i \times h^i \times w^i}$ is the feature at the $i$-th layer of $E_c$ and $L$ is the total number of layers. 
The context encoder resembles the U-Net encoder in its design, albeit shallower structure. Given the varying appearance of each frame caused by object motion, it is necessary to adjust the global context accordingly for effective prediction.
Inspired by the success of text-image fusion using affine transformation \cite{tao2020df}, we propose motion-conditioned affine transformation to refine the global context for each frame. We first obtain the motion feature $M_{t+1} \in \mathbb{R}^{2c}$ by concatenating the output of the motion encoder $E_m$ and audio encoder $E_a$ as follow:
\begin{equation}
    M_{t+1} = E_m\left(V_t, \hat{F}_{t+1}\right) || E_a\left(A_t\right),
\end{equation}
where $||$ denotes the concatenation operation. We also employ audio to compute $M_{t+1}$ as it contains motion-related information. 
The transformation parameters $\gamma_{t+1}^i \in \mathbb{R}^c$ and $\beta_{t+1}^i \in \mathbb{R}^c$ for the global context feature at the $i$-th layer are obtained from two separate MLPs, which are both conditioned on motion feature $M_{t+1}$ so that the transformed feature is able to adapt to motion variance:
\begin{equation}
    \gamma_{t+1}^i = \operatorname{MLP}_1^i\left(M_{t+1}\right), \quad    \beta_{t+1}^i = \operatorname{MLP}_2^i\left(M_{t+1}\right).
\end{equation} 
Then, channel-wise scaling and shifting is performed on $\text{z}^i$ to obtain the adapted context $\text{z}_{t+1}^i \in \mathbb{R}^{c^i \times h^i \times w^i} $:
\begin{equation}
    \text{z}_{t+1}^i = \gamma_{t+1}^i \cdot \text{z}^i + \beta_{t+1}^i.
\end{equation}
Finally, we insert $\text{z}_{t+1}^i$ into its corresponding layer in the U-Net decoder.

\subsection{Optimization}
We train our model in two stages due to the different convergence speeds of MME and CAR. We optimize MME in the first stage, and then we fix MME and optimize CAR in the second stage. We train MME by optimizing an optical flow reconstruction loss $\mathcal{L}_{\text{flow}}$ and a smoothness regularization term $\mathcal{L}_{\text{smooth}}$:
\begin{equation}
    \small \mathcal{L}_{\text{MME}} = \mathcal{L}_\text{flow} + \lambda_{\text{smooth}}\mathcal{L}_{\text{smooth}},
\end{equation}
where $\lambda_{\text{smooth}}$ is 
a weighting hyperparameter for $\mathcal{L}_{\text{smooth}}$. $\mathcal{L}_{\text{flow}}$ computes a mean-squared error between the predicted optical flow and ground truth optical flow as follows:
\begin{equation}
    \mathcal{L}_{\text{flow}} = \sum_{t=K+1}^{T}\left\|F_{t}-\hat{F}_{t}\right\|_{2}^{2}.
\end{equation}
Following \cite{wu2020future,yin2018geonet}, we have a smoothness regularization term $\mathcal{L}_{\text{smooth}}$ for our optical flow estimation:
\begin{equation}
\small \mathcal{L_{\text {smooth }}}=\sum_{t=K+1}^{T}\left\|\nabla \hat{F}_{t}\right\|_{1} e^{-\left\|\nabla V_{t}\right\|_{1}},
\end{equation}
where $\nabla$ is the gradient operator. 
The supervision for CAR is an image  reconstruction loss $\mathcal{L}_v$, which computes the mean-squared error between the predicted and ground truth frames:
\begin{equation}
   \small \mathcal{L}_v = \sum_{t=K+1}^{T}\left\|V_{t}-\hat{V}_{t}\right\|_{2}^{2}.
\end{equation}

\section{Experiments}
\begin{table*}[t]
\centering
\resizebox{\linewidth}{!}{
    \begin{tabular}{l|p{1cm}<\centering|p{1.5cm}<\centering p{1.5cm}<\centering p{1.5cm}<\centering p{1.5cm}<\centering|p{1.5cm}<\centering p{1.5cm}<\centering p{1.5cm}<\centering p{1.5cm}<\centering}
        \toprule 
        \multirow{2}{*}{Method} & \multirow{2}{*}{Type} & \multicolumn{4}{c|}{ SSIM $\uparrow$} & \multicolumn{4}{c}{ PSNR $\uparrow$} \\
         & & Fr 6 & Fr 15 & Fr 25 & Mean & Fr 6 & Fr 15 & Fr 25 & Mean \\
        \midrule
        Denton and Fergus \cite{denton2018stochastic} & V & $0.9265$ & $0.8300$ & $0.7999$ & $-$ &  $18.59$ & $14.65$ & $13.98$ & $-$\\
        MSPred \cite{villar2022MSPred} & V & 0.9400 & 0.8903 & 0.8060 & 0.8846 & 21.57 & 19.02 & 17.04 & 19.08 \\
        \midrule
        Vougioukas \etal \cite{vougioukas2018end} & M & $0.8600$ & $0.8571$ & $0.8573$ & $-$ & $15.17$ & $14.99$ & $15.01$ & $-$\\
        Sound2Sight \cite{chatterjee2020sound2sight} & M & $0.9505$ & $ 0.8780$ & $0.8749$ & $0.8910$ & $22.22$ & $ 18.16$ & $17.84$ & $18.70$ \\
        Our Method & M & $\textbf{0.9608}$ & $\textbf{0.9158}$ & $\textbf{0.8990}$ & $\textbf{0.9195}$ & $\textbf{23.61}$ & $\textbf{19.88}$ & $\textbf{18.99}$ & $\textbf{20.23}$ \\ 
        \bottomrule
    \end{tabular}
    }
    \vspace{-3mm}
    \caption{Comparison on Multimodal MovingMNIST with 5 seen frames. Higher values are better for SSIM and PSNR. The best results are marked in bold. Type M denotes method using audio-visual modality and Type V denotes only using visual modality. 
}
\label{tbl1}
\vspace{-3mm}
\end{table*}

\begin{figure*}[t]
\centering
\includegraphics[scale=0.42]{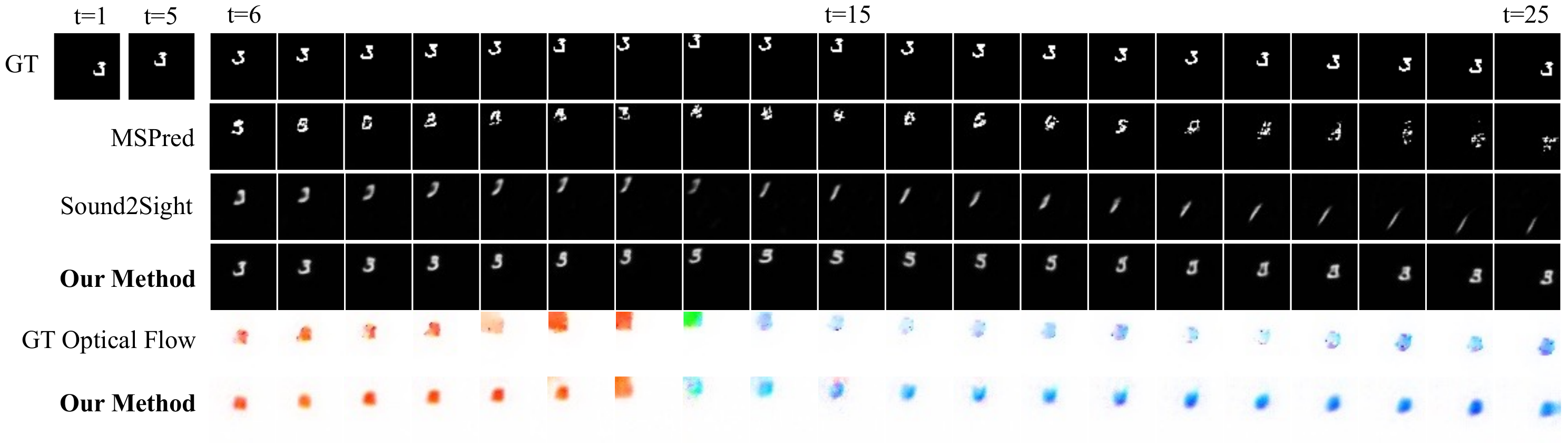}
\caption{Qualitative results on Multimodal MovingMNIST.}
\label{mnist-img}
\end{figure*}

\noindent \textbf{Datasets.}
We conduct experiments on both synthetic and real-world datasets taken from \cite{chatterjee2020sound2sight}:
1) \textbf{Multimodal MovingMNIST.} This dataset is an extension of stochastic MovingMNIST \cite{srivastava2015unsupervised} by adding artificial sound. Although MovingMNIST is synthetic, it is a standard benchmark in video prediction \cite{wang2018eidetic, lee2021video, denton2018stochastic}.  In Multimodal MovingMNIST, each digit is equipped with a unique tone and the amplitude of the tone is inversely proportional to the distance of the digit from the origin. The sound changes momentarily whenever the digit hits the boundary. This dataset consists of 8,000 training and 1,000 test videos.
2) \textbf{YouTube Painting.} It contains painting videos from YouTube. In each video, a painter is painting on a canvas in an indoor environment and there is a clear sound of the brush strokes. It contains $4.8\mathrm{~K}$ training, and 500 test videos. Each video is of 64 $\times$ 64 resolution, 30fps, and 3 seconds long.
3) \textbf{AudioSet-Drums.} This dataset contains selected videos from the drums class of AudioSet \cite{gemmeke2017audio}. The video clips are selected such that the drum player is visible when the drum beat is audible. It contains $6\mathrm{~K}$ training, and $1\mathrm{~K}$ test videos. Each video is of 64 × 64 resolution, 30fps, and 3 seconds long.

\vspace{1mm}
\noindent \textbf{Implementation.}
The video frames are resized to $48 \times 48$ in Multimodal MovingMNIST and $64 \times 64$ in YouTube Painting and AudioSet-Drums. The intensity values of the video frames are normalized into $[0, 1]$. We directly use the precomputed spectrogram by \cite{chatterjee2020sound2sight} as audio input. We use the Adam optimizer \cite{kingma2014adam} and the learning rate is set to $1e-4$ and $1e-3$ for MME and CAR, respectively. The coefficient $\lambda_{\text{smooth}}$ is set to 0.01.
We use 4-layer ConvLSTMs in MME and insert $L=4$ layers of context feature in CAR.
Ground truth optical flow is precomputed using the TV-L1 algorithm from \cite{zach2007duality}. We implement our framework with Pytorch on two Nvidia GeForce GTX1080Ti GPUs.

\vspace{1mm}
\noindent \textbf{Evaluation setup.}
On Multimodal MovingMNIST, we show 5 initial frames and full audio, and we predict the next 15 frames during training and 20 frames during evaluation. On YouTube Painting and AudioSet-Drums, we show 15 initial visual frames and full audio, and predict the next 15 visual frames during training and 30 visual frames during evaluation. We evaluate our method by the structure similarity (SSIM) \cite{wang2004image} and Peak Signal to Noise Ratio (PSNR). We report SSIM and PSNR on selected visual frames. The mean SSIM and PSNR of the full predicted sequence are also reported to represent average performance. 
\subsection{Results}
\begin{wrapfigure}{r}{0.6\textwidth}
\vspace{-3mm}
\centering
\includegraphics[scale=0.46]{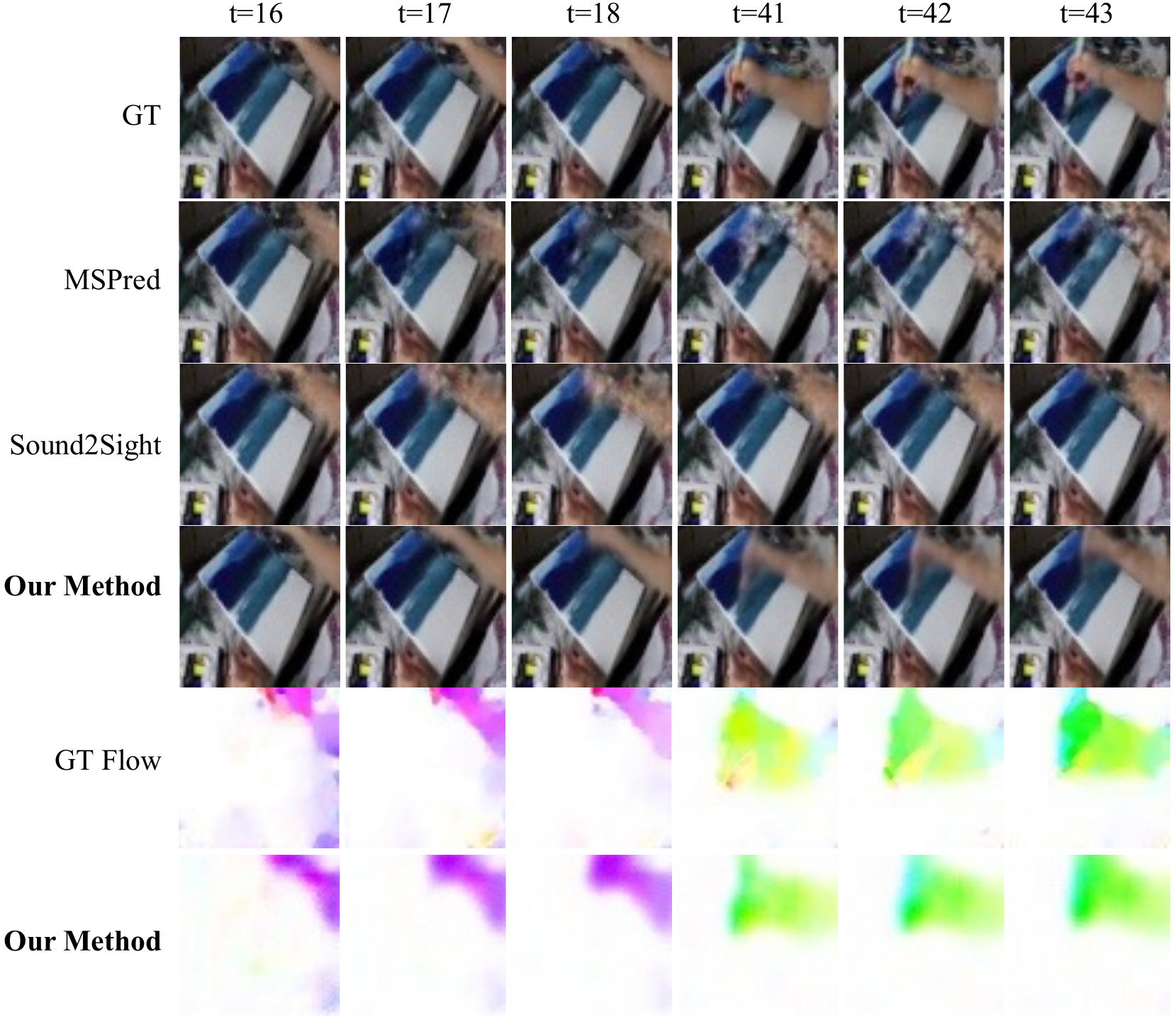}
\vspace{-8mm}
\caption{Qualitative results on YouTube Painting.}
\label{youtube-img}
\vspace{-3mm}
\end{wrapfigure}
We compare our results with Sound2Sight \cite{chatterjee2020sound2sight}, Vougioukas \etal \cite{vougioukas2018end}, MSPred \cite{villar2022MSPred} and Denton and Fergu \cite{denton2018stochastic}. 
We re-run the source code of Sound2Sight \cite{chatterjee2020sound2sight} to compute the average score over 100 times of sampling instead of using the best result of the samples drawn from sampling reported by Sound2sight.
In practice, it is statistically more meaningful to use the mean instead of the best result since the stochastic approach cannot always achieve the highest result in every sampling. Moreover, we do not know the ground truths of future visual frames. Therefore, it is unreasonable for \cite{chatterjee2020sound2sight} to search for the best samples drawn from their stochastic model using the ground truths.

\begin{table*}[t]
\centering
\resizebox{1\linewidth}{!}{
	\begin{tabular}{l|p{1cm}<\centering|p{1.5cm}<\centering p{1.5cm}<\centering p{1.5cm}<\centering p{1.5cm}<\centering|p{1.5cm}<\centering p{1.5cm}<\centering p{1.5cm}<\centering p{1.5cm}<\centering}
        \toprule 
        \multirow{2}{*}{Method} & \multirow{2}{*}{Type} & \multicolumn{4}{c|}{ SSIM $\uparrow$ } & \multicolumn{4}{c}{ PSNR $\uparrow$ } \\
         & & Fr 16 & Fr 30 & Fr 45 & Mean & Fr 16 & Fr 30 & Fr 45 & Mean \\
        \midrule
        Denton and Fergus \cite{denton2018stochastic} & V & $0.9779$ & $0.6654$ & $0.4193$ & $-$ &  $32.52$ & $16.05$ & $11.84$ & $-$\\
        MSPred \cite{villar2022MSPred} & V & 0.9648 & 0.8991 & 0.8617 & 0.8965 & 33.42 & 26.42 & 24.25 & 26.65 \\
        \midrule
        Vougioukas \etal \cite{vougioukas2018end} & M & $0.9281$ & $0.9126$ & $0.9027$ & $-$ & $26.97$ & $25.58$ & $24.78$ & $-$\\
        Sound2Sight \cite{chatterjee2020sound2sight} & M & $ 0.9716$ & $0.9261$ & $0.9074
        $ & $ 0.9264
        $ & $ 31.91
        $ & $26.73
        $ & $ 25.17
        $ & $ 26.95
        $ \\
        Our Method &M& $\textbf{0.9848}$ & $\textbf{0.9284}$ & $\textbf{0.9104}$ & $\textbf{0.9313}$ & $\textbf{35.12}$ & $\textbf{27.19}$ &$\textbf{25.53}$ & $\textbf{27.70}$ \\
        \bottomrule
    \end{tabular}
    }
\vspace{-3mm}
\caption{Comparison on YouTube Painting with 15 seen frames.
}	
 \label{tbl2}
 \vspace{-3mm}
\end{table*}
\begin{table*}[t]
\centering
\resizebox{\linewidth}{!}{
\begin{tabular}{l|p{1cm}<\centering|p{1.5cm}<\centering p{1.5cm}<\centering p{1.5cm}<\centering p{1.5cm}<\centering|p{1.5cm}<\centering p{1.5cm}<\centering p{1.5cm}<\centering p{1.5cm}<\centering}
    \toprule 
    \multirow{2}{*}{Method} & \multirow{2}{*}{Type} & \multicolumn{4}{c|}{ SSIM  $\uparrow$ } & \multicolumn{4}{c}{ PSNR  $\uparrow$ } \\
     & & Fr 16 & Fr 30 & Fr 45 & Mean & Fr 16 & Fr 30 & Fr 45 & Mean \\
    \midrule
    Denton and Fergus \cite{denton2018stochastic} & V & $0.9706$ & $0.6606$ & $0.5097$ & $-$ &  $30.01$ & $16.57$ & $13.49$ & $-$\\
    MSPred \cite{villar2022MSPred} & V & 0.9799 & 0.9382 & 0.9214 & 0.9389 & 33.55 & 27.30 & 25.91 & 27.60 \\
    \midrule
    Vougioukas \etal \cite{vougioukas2018end} & M & $0.8986$ & $0.8905$ & $0.8866$ & $-$ & $23.62$ & $23.14$ & $22.91$ & $-$\\
    Sound2Sight \cite{chatterjee2020sound2sight} & M & $0.9875
    $ & $0.9524
    $ & $0.9434
    $ & $0.9544
    $ & $34.23
    $ & $27.73
    $ & $\textbf{26.71}
    $ & $28.13
    $ \\
    Our Method &M& $\textbf{0.9896}$ & $\textbf{0.9533}$ & $\textbf{0.9437}$ & $\textbf{0.9558}$ & $\textbf{35.00}$ & $\textbf{27.76}$ &$26.68$ & $\textbf{28.22}$ \\
    \bottomrule
\end{tabular}
}
\vspace{-3mm}
 \caption{Comparison on AudioSet-Drums with 15 seen frames.} 
\label{tbl3}
\vspace{-2mm}
\end{table*}

\vspace{1mm}
\noindent \textbf{Multimodal MovingMNIST.}
Tab.~\ref{tbl1} shows quantitative results on Multimodal MovingMNIST. Our method outperforms others by a large margin across all time steps. Fig.~\ref{mnist-img} shows the quantitative results. The first four rows show the comparison of the predicted visual frames sequence. The last two rows show the corresponding optical flow results. We follow the flow color coding of \cite{baker2011database}. It can be seen that our method is able to correctly predict the trajectory and also maintain the complex shape of digit `3' in long-term prediction. On the contrary, the predicted sequence from Sound2Sight and MSPred gradually lose the shape of the digit or even deviate from the correct trajectory.

\vspace{1mm}
\noindent \textbf{YouTube Painting.}
Tab.~\ref{tbl2} shows quantitative results on YouTube Painting.  Our model again surpasses other methods.
Fig.~\ref{youtube-img} shows the qualitative comparison with Sound2Sight. Due to entangled modeling of motion and appearance, the result of Sound2Sight gradually loses the shape of the hand. Furthermore, the hand becomes stationary at the top right corner of the image. In contrast, our model maintains a relatively good appearance and generally follows the movement of the hand in ground truth frames. We can also see that our predicted optical flows are close to the ground truths.

\vspace{1mm}
\noindent \textbf{AudioSet-Drums.}
Tab.~\ref{tbl3} shows quantitative results on AudioSet-Drums. Our model surpasses others at almost every time step. 
We notice that our superiority is not as obvious on AudioSet-Drums as on the other two datasets. The reason may be that motion only exists in a small region of frames in AudioSet-Drums, and is hard for our model to fully capture it. Nevertheless, our model still achieves the best overall result.

\begin{table}[t]
\begin{minipage}{0.46\linewidth}
\centering
\setlength{\tabcolsep}{1pt}
\resizebox{\linewidth}{!}{
\begin{tabular}{l|p{2cm}<\centering p{2cm}<\centering p{2cm}<\centering}
\toprule 
Method & YouTube & MNIST & AudioSet \\
\midrule
V & $11.25$ & $-$ & $-$ \\
V + Recall & $10.43$ &$6.40$ & $4.63$ \\
MME & $\textbf{9.85}$ & $\textbf{4.82}$ & $\textbf{3.97}$ \\
\bottomrule
\end{tabular}
}
\vspace{-3mm}
\caption{Analysis of audio in MME. AEPE results are presented in $10^{-2}$ scale.}
\label{ablation-MME}
\end{minipage}
\hspace{0.4cm}
\begin{minipage}{0.46\linewidth}
\centering
\setlength{\tabcolsep}{1pt}
\resizebox{\linewidth}{!}{
\begin{tabular}{l|p{1.4cm}<\centering p{1.4cm}<\centering p{1.4cm}<\centering p{1.4cm}<\centering}
    \toprule
    \multirow{2}{*}{ Method } & \multicolumn{4}{c}{ SSIM $\uparrow$} \\
    & Fr 16 & Fr 30 & Fr 45 & Mean \\
    \midrule 
    MME+Unet & $\textbf{0.9873}$ & $0.9105$ & $0.8761$ & $0.9143$ \\
    MME+Unet+ContextEnc & $0.9821$ & $0.9266$ & $ 0.9091$ & $0.9291$ \\
    MME+CAR &$0.9848$ & $\textbf{0.9284}$ & $\textbf{0.9104}$ & $\textbf{0.9313}$ \\
    \bottomrule
\end{tabular}
}
\vspace{-3mm}
\caption{Analysis of CAR on YouTube Painting.}
\label{ablation-CAR}
\end{minipage}
\vspace{-5mm}
\end{table}

\vspace{-3mm}
\subsection{Ablation Study}
\noindent \textbf{Effectiveness of audio-motion correlation in MME.} 
As audio is directly influencing optical flow prediction, we analyze the effectiveness of audio in MME. We compare the quality of optical flow using average endpoint error (AEPE) \cite{sun2018pwc,ilg2017flownet} in Tab.~\ref{ablation-MME}. The baseline `V' predicts optical flow without audio. `V+Recall' \textit{Recall} from audio motion memory, but does not condense it. `MME' is our full model in stage 1 with both \textit{condense} and \textit{recall}. MME has the lowest error in the flow prediction, suggesting the audio is effective in helping the visual motion prediction.

\vspace{1mm}
\noindent \textbf{Effectiveness of Context-Aware Refinement.}
We analyze context-aware refinement (CAR) in Tab.~\ref{ablation-CAR}. `MME+Unet' only uses U-Net for refinement in stage 2. `MME+Unet+ContextEnc' adds a context encoder but does not perform affine transformation over the context feature. 
`MME+CAR' is our full model with motion-conditioned affine transformation. We can see from the first and second rows that a single U-Net is insufficient to model image appearance in long-term prediction, and thus it is very important to supply additional context information. 
We can see from the second and last rows that our motion-conditioned affine transformation is able to adjust the global context feature for better performance in long-term predictions. 

\begin{wraptable}{R}{0.45\textwidth}
\centering
\vspace{-2mm}
\resizebox{\linewidth}{!}{
\begin{tabular}{c|cccc}
\hline
Method          & Fr 16       & Fr 30       & Fr 45       & Mean       \\ \hline
Sound2Sight \cite{chatterjee2020sound2sight} & 0.9556      & 0.9107      & 0.8936      & 0.9117     \\
Ours    & \textbf{0.9827}      & \textbf{0.9252}      & \textbf{0.9069}      & \textbf{0.9280}     \\ \hline
\end{tabular}
}
\vspace{-3mm}
\caption{SSIM results on YouTube Painting with $96 \times 96$ videos.}
\label{96}
\vspace{-4mm}
\end{wraptable}

\vspace{1mm}
\noindent \textbf{Input video with higher resolution.}
\ytComment{
We use the video frames of original resolution $96 \times 96$ as input. The results are shown in Tab. ~\ref{96}. Our model consistently outperforms \cite{chatterjee2020sound2sight}, demonstrating its effectiveness even when provided with higher-resolution videos.
}

\section{Conclusions}
\label{sec:conclusion}
In this paper, we address the task of audio-visual conditioned video prediction by decoupling it into motion and appearance. We design the MME and CAR by considering the influence of audio and visual modalities in each component. MME leverages audio-motion correlation by using audio as an external motion memory in predicting future optical flow. CAR addresses the loss of appearance context in long-term prediction by adding a context encoder. To make the global context feature compatible with motion variance, we perform the motion-conditioned affine transformation to adjust the feature. The effectiveness of our proposed method is analyzed both quantitatively and qualitatively.

\paragraph{Acknowledgement.}This research is supported by the National Research Foundation, Singapore under its AI Singapore Programme (AISG Award No: AISG2-RP-2021-024),
and the Tier 2 grant MOE-T2EP20120-0011 from the Singapore Ministry of Education.

\appendix
\renewcommand\thefigure{\Alph{section}\arabic{figure}}  
\renewcommand\thetable{\Alph{section}\arabic{table}}

{\centering\section*{Supplementary Material}}

\section{Qualitative Ablation Study}
\setcounter{figure}{0}
\setcounter{table}{0}

We conduct qualitative analysis to verify the effectiveness of MME and CAR.

\begin{figure}[h]
\centering
\includegraphics[scale=0.53]{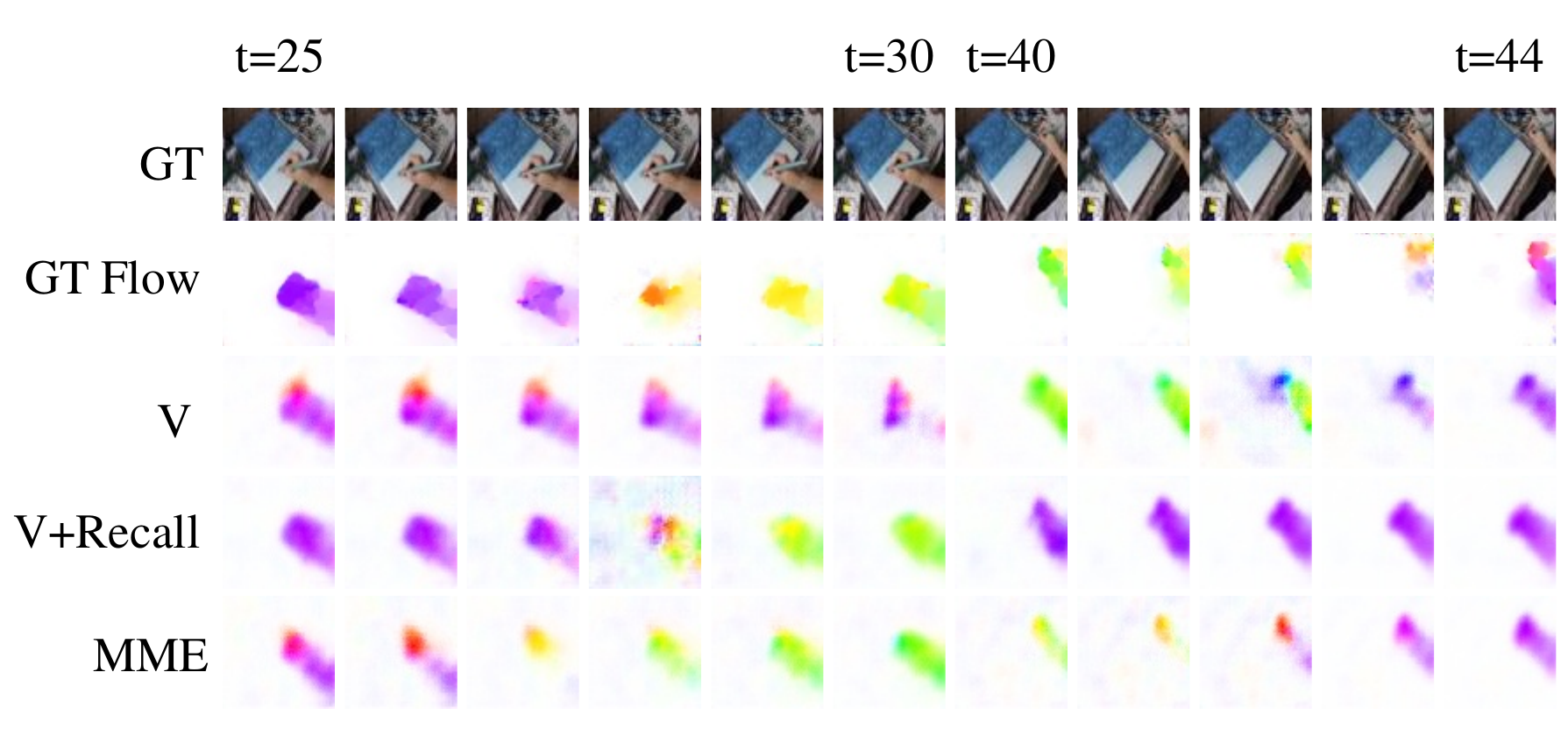}
\caption{Qualitative comparison of different motion estimation networks.}
\label{ablation-mme-img}
\end{figure}

\noindent \textbf{Effectiveness of audio-motion correlation in MME.} Fig.~\ref{ablation-mme-img} presents the qualitative comparion of different motion estimation networks. `GT Flow' is the ground truth optical flow.
The baseline `V' predicts optical flow without audio. `V+Recall' \textit{Recall} from audio motion memory, but does not condense it. `MME' is our full model in stage 1 with both \textit{condense} and \textit{recall}. It is clear to see that only using visual modality (`V') to predict motion has the worst prediction.
Although `V+Recall' captures the correct motion in the short term, it fails at longer future due to its large-size motion memory. In contrast, MME has a better balance in both short-term and long-term predictions due to the condensed memory.

\begin{figure}[h]
\centering
\includegraphics[scale=0.38]{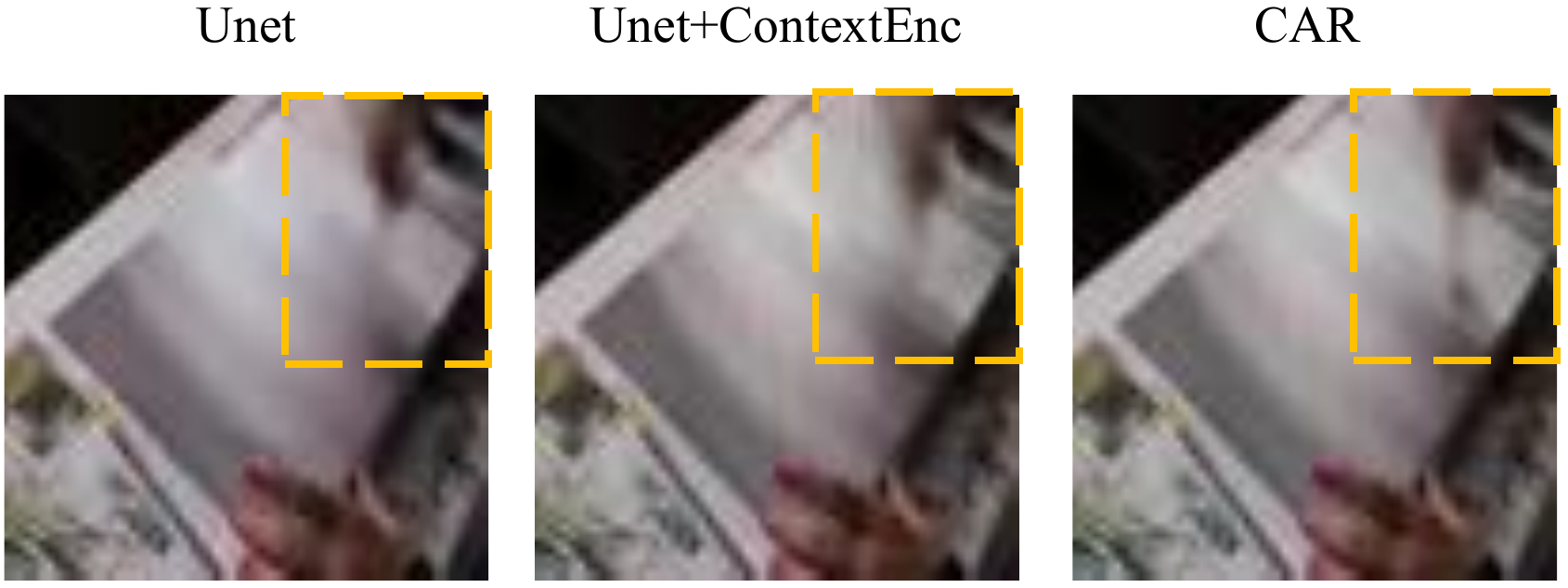}
\vspace{1mm}
\caption{Qualitative comparison of different refinement networks.}
\label{ablation-CAR-img}
\vspace{3mm}
\end{figure}

\noindent \textbf{Effectiveness of Context-Aware Refinement.}
Fig.~\ref{ablation-CAR-img} presents the qualitative comparison of different image refinement networks.
`Unet' only uses U-Net for refinement in stage 2. `Unet+ContextEnc' adds a context encoder but does not perform affine transformation over the context feature. It shows that CAR is able to provide and aggregate the feature of hand and pen to make the moving object more concrete. 

\section{Qualitative Video Results}
We provide a video (`video results.mp4') to show the prediction by our model and the ground truth in the supplementary zip file. The white denotes ground truth and the red denotes prediction. Video 1 and video 2 show 15 past frames and 30 future frames. Video 3 and 4 show 5 past frames and 20 future frames. Our prediction is temporally smooth and consistent with the ground truth audio and visual frames.

\section{Architecture Details}
\label{sec:architecture}
We present the detailed architecture of our proposed framework. 

\vspace{2mm}
\noindent \textbf{Motion Encoder.} The motion encoder consists of four convolutional layers with kernel size of $4 \times 4$ and stride of 2. The number of channels is $\left\{64, 64, 128, 128\right\}$. Each convolutional layer is followed by Batch Normalization and Leaky ReLU, except for the last layer which is followed by Batch Normalization and Tanh.

\vspace{2mm}
\noindent \textbf{Motion Decoder.}
The motion decoder consists of four deconvolution layers with kernel size of $4 \times 4$ and stride of 2. The number of channels is $\left\{128, 64, 64, 2\right\}$. Each convolutional layer is followed by Batch Normalization and Leaky ReLU, except for the last layer.

\vspace{2mm}
\noindent \textbf{Audio Encoder.}
The audio encoder consists of five convolutional layers with kernel size of $4 \times 4$ for the first four layers and with kernel size of $2 \times 2$ for the last layer. The number of channels is $\left\{64, 128, 256, 512, 128\right\}$. Each convolutional layer is followed by Batch Normalization and Leaky ReLU,  except for the last layer which is followed by Batch Normalization and Tanh.

\begin{figure}[h]
\centering
\setlength{\abovecaptionskip}{0.1cm}
\includegraphics[scale=0.6]{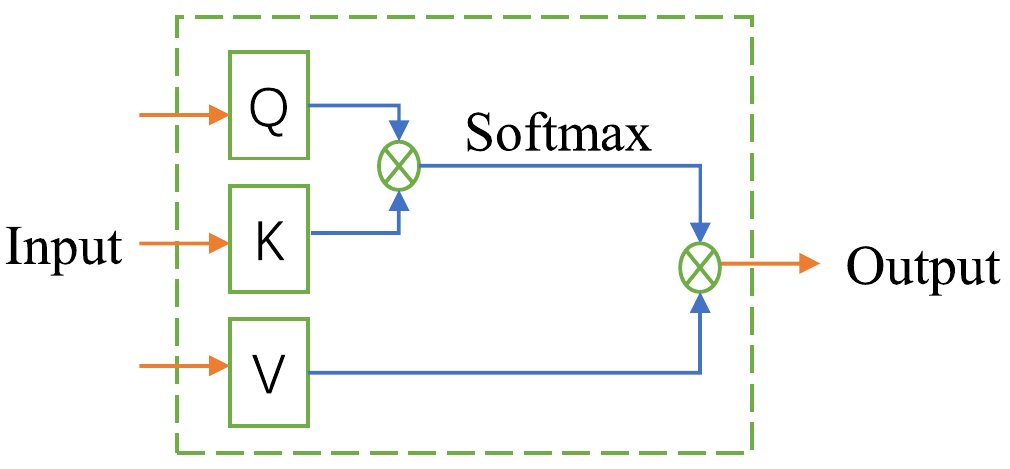}
\caption{Illustration of the attention mechanism.}
\label{attention}
\vspace{-3mm}
\end{figure}

\vspace{2mm}
\noindent \textbf{Condense and Recall.}
The operators $\operatorname{Condense}$ and $\operatorname{Recall}$ use attention mechanism as shown in Fig~\ref{attention}. Specifically, Operator $\operatorname{Condense}$ takes the motion memory $\operatorname{MM}$ as input and projects it into the query $(\mathbf{Q})$, key $(\mathbf{K})$ and value $(\mathbf{V})$ via three separate 1D convolutional layers with kernel size of 1, respectively. In $\operatorname{Recall}$, the visual feature is sent into a 2D convolutional layer with kernel size of $1 \times 1$ and then is flattened along spatial dimensions as $\mathbf{Q}$. The condensed memory is sent into two separate 1D convolutional layers with kernel size of 1 and projected as $\mathbf{K}$ and $\mathbf{V}$,  respectively.

\vspace{2mm}
\noindent \textbf{Context Encoder.}
The context encoder has four blocks, where each block consists of two convolutional layers followed by Batch Normalization and ReLU. A max pooling layer is inserted between adjacent blocks.


\bibliography{bmvc_review}
\end{document}